\documentclass{article}

\usepackage[final]{nips_2017}

\usepackage[utf8]{inputenc} 
\usepackage[T1]{fontenc}    
\usepackage{hyperref}       
\usepackage{url}            
\usepackage{booktabs}       
\usepackage{amsfonts}       
\usepackage{nicefrac}       
\usepackage{microtype}      
\usepackage[pdftex]{graphicx}
\usepackage{caption}
\usepackage{subcaption}
\usepackage{nth}
\usepackage{amsmath,amsthm,mathtools,dsfont}
\usepackage{ textcomp }

\title{Probabilistic Generative Adversarial Networks}

\author{
  Hamid Eghbal-zadeh\thanks{This work was supported by the Austrian Ministries
BMVIT and BMWFW, and the Province of Upper Austria,
via the COMET Center SCCH. We also gratefully acknowledge
the support of NVIDIA Corporation with the
donation of a Titan X GPU used for this research} \\
  Department of Computational Perception\\
Johannes Kepler University\\
   Linz, Austria\\
  \texttt{hamid.eghbal-zadeh@jku.at} \\
 \And
  Gerhard Widmer \\
  Department of Computational Perception\\
Johannes Kepler University\\
   Linz, Austria\\
  \texttt{gerhard.widmer@jku.at} \\
}

\begin{document}

\maketitle

\begin{abstract}
We introduce the Probabilistic Generative Adversarial Network (PGAN),
a new GAN variant based on a new kind of objective function.
The central idea is to integrate a probabilistic model (a Gaussian
Mixture Model, in our case) into the GAN framework which supports
a new kind of loss function (based on likelihood rather than classification
loss),  and at the same time gives a meaningful measure of the quality of
the outputs generated by the network.
Experiments with MNIST show that the model learns to generate
realistic images, and at the same time computes likelihoods that are
correlated with the quality of the generated images.
We show that PGAN is better able to cope with instability problems
that are usually observed in the GAN training procedure.
We investigate this from three aspects: the probability landscape of
the discriminator, gradients of the generator,
and the \emph{perfect discriminator} problem.
\end{abstract}

\section{Introduction}

Generative Adversarial Networks (GANs) \citep{goodfellow2014generative} are considered by many one of the greatest advances in deep learning domain during recent years.
GANs are able to learn the Probability Density Function (PDF) underlying a set of real images and, based on this, can learn to generate realistic and sharp new images.
Many different variations on GANs have already been proposed, but some issues still remained unsolved.

One of the main issues in GANs is the lack of a meaningful measure for quantifying the quality of a GAN's output.
On the one hand this issue makes the training of GANs more complicated and prevents us from detecting a possible training degradation.
On the other hand, it makes the comparison of different GANs and different architectures very difficult and also dependent on an intuitive visual assessment of the generated images.
This is in contrast to, e.g., Variational Auto-Encoders (VAEs), where the quality of a model can be assessed by the likelihood of the model.
However, VAEs are known to be unable to generate sharp images, and the images generated with VAEs tend to be blurry because of the injected noise and the imperfect reconstruction.

Another known issue with GANs is the instable training where either the discriminator or the generator may take over the other and the model cannot recover from that.
This issue has been well investigated in \citep{arjovsky2017towards}, where tools are provided to investigate the sources of this unstable training behaviour.

In this paper, we introduce a new member of the GAN family, the Probabilistic GAN (PGAN).
By integrating a Gaussian Mixture Model (GMM) in the discriminator, we introduce a GMM likelihood 
loss function (rather than a loss based on classification error) for optimizing the generator and discriminator in the GAN minimax game.
The advantages of this are (a) having a probabilistic loss in terms of a Gaussian likelihood which is strongly correlated with the quality of the generated images; (b) higher training stability due to not relying on the fake images in the discriminator and only focusing on the distribution of the real images; and (c) having an On\textbackslash Off mechanism for the gradients of the generator and discriminator when the generator does not improve; this way, the gradients received in the generator are more suitable for the GAN training updates.
We will demonstrate these advantages by providing empirical results, showing that likelihood and the generated images correlate, that
the network learns a probability landscape over feature space that
makes it more robust against \emph{mode dropping}, and that, when the generator is fixed and the discriminator `takes over', the gradients in the generator become small, and when the generator is updating once again, the gradient norm returns to its  normal scale (cf. 
\citep{arjovsky2017towards}).


\section{Related Work}
\label{sec:related}
GANs
have advanced the state of the art in deep generative models by enabling a better PDF estimation via neural networks, which can be used, e.g., for image generation.
While they exhibit a surprising ability to generate sharp and realistic images -- in contrast the blurry images generated via other techniques such as VAEs --, they are known to be difficult to train.

Since the advent of the original GAN, many variations have been proposed to make training easier.
The Least Squares GAN (LSGAN) \citep{mao2016least} reformulates the original loss function into
a quadratic least squares form which drastically increases training robustness.
The Energy-Based GAN (EBGAN) \citep{zhao2016energy} uses a reconstruction loss as an energy function that assigns 
low energies to the regions near the data manifold and higher energies to others.
The recently introduced Wasserstein GAN (WGAN) \cite{arjovsky2017wasserstein} use the Wasserstein distance (also known as Earth Mover's Distance) to provide a meaningful distance measure which correlates with the quality of the generated images.
\cite{arjovsky2017towards} investigate the sources of instability in GAN training, and introduce new tools to spot the sources of instability and saturation in training GANs.

Similar to WGAN, our PGAN provides a meaningful loss function; in our case, this is a Gaussian likelihood relative to generated images.
For more robust training, we build this likelihood into the loss function of the generator and the discriminator, in a least squares form similar to LSGAN.
To demonstrate the behavior of PGAN during training, we use the tools introduced in \citep{arjovsky2017towards}, analyzing the gradients of the generator and the discriminator during training.
In addition, we set up an experiment to investigate the ability of PGAN to recover from an instable training mode when we prevent one of the players (generator/discriminator) from adapting and let the other take over.
In these experiments, we then investigate the gradient statistics to obtain a deeper understanding of the robustness of PGAN in critical training situations.

\section{GANs with Gaussian Density Approximation }
\label{sec:pgan}
\begin{figure}[h]
  \centering
  \fbox{\includegraphics[width=1.\linewidth]{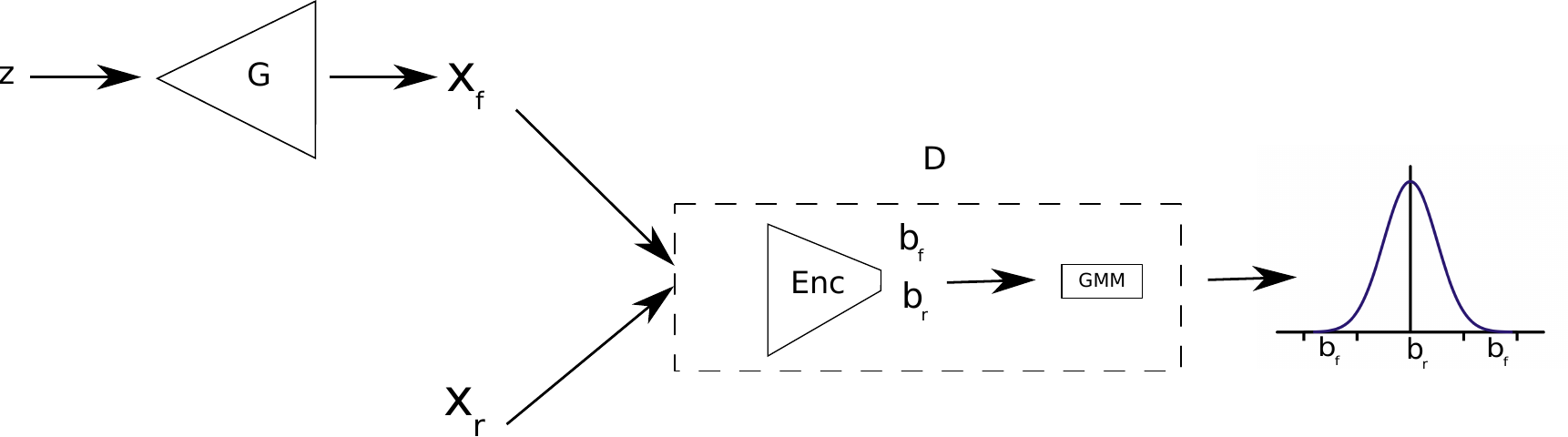}}
  \caption{Block diagram of PGAN. Discriminator $D$ consists of an encoder and a GMM. 
  The discriminator is trained in two steps: first, the GMM parameters are estimated from the encodings of the real images. Second, the encoder updates such that the likelihoods of 
  encoded fake images ($b_f$) are close to zero and likelihoods of encoded real images ($b_r$) are close to 1. 
 Intuitively, the encoder wants to push the $b_f$s to the side of the Gaussian bell, while keeping the $b_r$s close to the center.} \label{fig:block_diagram}
\end{figure}

In this section, we explain the main idea in PGAN and describe the training procedure of our discriminator.
As can be seen in Figure \ref{fig:block_diagram}, PGAN consists of a generator $G$ and a discriminator $D$.
The generator $D$ itself consists of two parts: 1) an encoder and 2) a Gaussian Mixture Model (GMM).
The encoder encodes images into a \emph{bottleneck embedding space}.
For an input image $\boldsymbol{x}_i$, the encoder creates a bottleneck embedding $\boldsymbol{b}_i$ which is simply the activations of the bottleneck layer for an input image.
A GMM is then fit to the bottleneck embeddings of the real images, to give an estimation of the probability distribution over real images (in bottleneck embedding space).
When a discrimination between real and fake images is needed, the encoder first encodes the input $\boldsymbol{x}_i$ into $\boldsymbol{b}_i$.
With the parameters of the GMM estimated, a Gaussian likelihood is calculated for the incoming embedding.
This likelihood will be interpreted as the probability of $\boldsymbol{b}_i$ being an embedding of a real image, given the current model.

Our generator $G$ is a regular neural network, decoding a random variable $z$ from a Gaussian random distribution into an image.

\subsection{The GAN Objective}
\label{subsec:gan_objective}

In GANs, the general scenario is that a discriminator $D$ trying to distinguish between real and generated (fake) images competes with a generator $G$ that tries to fool the discriminator $D$ by generating realistic images.
The loss of the discriminator $D$ in the original GAN
\citep{goodfellow2014generative} is defined as:
\begin{equation}
\label{eq:disc_loss}
\mathcal{L}_{D} = \mathbb{E}_{\boldsymbol{x} \sim P_{\mathit{real}} } \big[ \log D(\boldsymbol{x}) \big] + \mathbb{E}_{\boldsymbol{z}\sim P_{\mathit{fake}} } \big[ \log (1 - D(G(\boldsymbol{z}))) \big] 
\end{equation}

(where it is assumed that for a given input image (vector) $\boldsymbol{x}$, the discriminator $D$ outputs a number -- the estimated probability of the image coming from the real set), and the loss of the generator $G$ is defined as:

\begin{equation}\label{eq:gen_loss}
\mathcal{L}_{G}=\mathbb{E}_{\boldsymbol{z} \sim P_{\mathit{fake}} } \big[ -\log D(G(\boldsymbol{z})) \big] 
\end{equation}

where
$\boldsymbol{z}$ is an observation from a random distribution $Z$; generator $G$ creates a fake image using this $\boldsymbol{z}$.

A modified version of the losses defined above using a least squares criterion is introduced in LSGAN, resulting in more stable training:

\begin{equation}\label{eq:ls_disc_loss}
{\mathcal{L}^{\mathit{ls}}_{D}}=\frac{1}{2}\mathbb{E}_{\boldsymbol{x} \sim P_{\mathit{real}} } \big[ (D(\boldsymbol{x})-1)^2 \big] + \frac{1}{2}\mathbb{E}_{\boldsymbol{z}\sim P_{\mathit{fake}} } \big[  (D(G(\boldsymbol{z}))-0)^2 \big] 
\end{equation}

and the loss of the generator $G$ is defined as:

\begin{equation}\label{eq:ls_gen_loss}
{\mathcal{L}^{\mathit{ls}}_{G}}=\frac{1}{2}\mathbb{E}_{\boldsymbol{z} \sim P_{\mathit{fake}} } \big[ ( D(G(\boldsymbol{z}))-1)^2 \big] 
\end{equation}

\subsection{The PGAN Objective}
\label{subsec:pgan_objective}

Similar to the other GAN discriminator objectives, the PGAN discriminator also tries to distinguish real from fake images, but only implicitly:
instead of drawing a boundary between real and fake samples, our discriminator fits a Gaussian on the embeddings of the projected real images. It then uses this GMM to provide a likelihood for both real and fake images.
A good discriminator, thus, would produce high probabilities for embeddings of real images and low probabilities for embeddings of fake images.
Therefore, we are mostly concerned with estimating the GMM parameters which would maximize our likelihood for the embeddings of the real images.

Denoting the encoding (embedding in bottleneck space) of an
image $\boldsymbol{x}$ as $\mathrm{enc}(\boldsymbol{x})$, we propose to define the discriminator loss (to be minimized) as follows:

\begin{equation}\label{eq:pgan_disc_loss}
{\mathcal{L}^{\mathit{pgan}}_{D}}=\frac{1}{2}\mathbb{E}_{\boldsymbol{x} \sim P_{\mathit{real}} } \big[ (\ell_\mathcal{M}(\boldsymbol{x})-1)^2 \big] \
+ \frac{1}{2}\mathbb{E}_{\boldsymbol{z}\sim P_{\mathit{fake}} } \big[  (\ell_\mathcal{M}(\mathrm{enc}(G(\boldsymbol{z})))-0)^2 \big] 
\end{equation}

and the loss of the generator $G$ as:

\begin{equation}\label{eq:pgan_gen_loss}
{\mathcal{L}^{\mathit{pgan}}_{G}}=\frac{1}{2}\mathbb{E}_{\boldsymbol{z} \sim P_{\mathit{fake}} } \big[ ( \ell_\mathcal{M}(\mathrm{enc}(G(\boldsymbol{z})))-1)^2 \big] 
\end{equation}

where the likelihood for a given image embedding $\boldsymbol{b} = \mathrm{enc}(\boldsymbol{x})$ is:
\begin{equation}\label{eq:mdn_bottleneck_loss}
\ell_\mathcal{M}(\boldsymbol{b}) = P(\boldsymbol{b} \mid \mathcal{M})
 = \sum_{i=1}^{K} w_{i} \cdot \mathcal{N} \Big( \boldsymbol{b}; \boldsymbol{\mu}_{i}, \boldsymbol{\Sigma}_{i} \Big)
\end{equation}

for a $K$-component GMM defined by parameters $w_i$ (mixture weights), $\boldsymbol{\mu}_i$ (mean vectors), and $\boldsymbol{\Sigma}_i$ (covariances).

\subsection{Iterative Training of the Discriminator}
\label{subsec:iterative}

In this section, we explain our iterative training procedure for the discriminator. 
In discriminator training, we use an iterative procedure where at first, we estimate the parameters of a GMM on the current bottleneck embeddings of the real images.
Given that, we update our likelihood function with the learned parameters and use this likelihood to update the parameters of the encoder.

Note again that in contrast to most of the discriminators used in GANs, we do not use a binary classifier.
We simply use the GMM likelihood in our objective function,
as a measure of the
probability of an image being from the distribution of the real images.
The discriminator objective in the second step will push the embeddings of the fake images to the side of the fitted GMM to achieve 
the lowest likelihood possible, since in the loss of the discriminator, such examples are punished by the second term in Eq. \ref{eq:pgan_disc_loss}.

This iterative two-step training procedure  can actually be seen as an Expectation Maximization (EM) algorithm as follows.
The first step in discriminator training is to estimate the parameters of the GMM using the bottleneck embeddings of \emph{only} the real data. 
This can be seen as \emph{The Expectation Step} which yields new probabilities over the embeddings and defines a new likelihood function.
The second step can be seen as a \emph{Maximization Step} where the parameters of the encoder are optimized to maximize the new likelihood function which was updated in the previous Expectation step.


As proven in\citep{shalev2014understanding}, in an EM algorithm the likelihood in each step can only increase compared to the previous step, assuming that the generator is fixed during the updates. 
In our case, given a fixed generator the GMM in the discriminator will represent the real embeddings better and better.
This we observe empirically in Figure \ref{fig:gen_nottrained}, where the likelihood increases quickly to 1 for the real data, and drops to 0 for the fake data.
In our implementation, we repeat this EM algorithm for each minibatch given to the discriminator.

\subsection{Gaussian Likelihood as a Meaningful Loss}
\label{subsec:meaningful_loss}

As the GMM is learned from the given real images, the likelihood produced using the model can naturally be interpreted as a measure of how well an arbitrary image fits into the distribution of our real images.
As a result, in each epoch this likelihood would explain how good the generated images are in terms of matching to the distribution of real images.
We will see an example of the correlation between perceived image
quality and likelihood in Section \ref{subsec:results} below.

\section{Empirical Results}
\label{sec:empirical_results}

The network architectures used in our experiment can be found in Table  \ref{tab:architectures} in Appendix\ref{ap:a}.
Generator and discriminator are a simplified version of DCGAN \citep{radford2015unsupervised}.
For our experiments, MNIST is used for training both an LSGAN and a PGAN.
Both networks are optimized with Adam \citep{kingma2014adam}, with
a learning rate of 0.0001 in both cases.
In the current experiments, a 1-dimensional bottleneck (i.e., 1 output unit only) and 1 Gaussian is used for PGAN.

\subsection{Likelihood Curves and Generated Images}
\label{subsec:results}
\begin{figure}[h]
  \centering
  \begin{subfigure}{.5\textwidth}
  \centering
  \includegraphics[width=1.\linewidth]{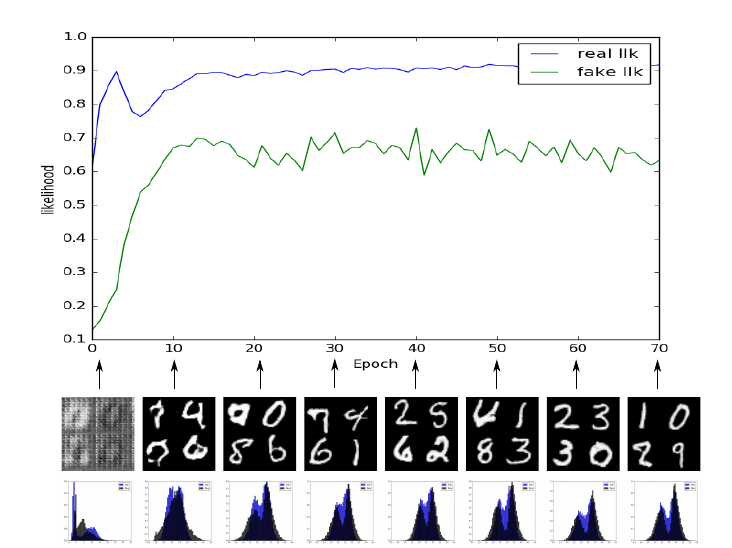}
  \caption{Generator and discriminator are trained. }
  \label{fig:gen_trained}
\end{subfigure}%
\begin{subfigure}{.5\textwidth}
  \centering
  \includegraphics[width=1.\linewidth]{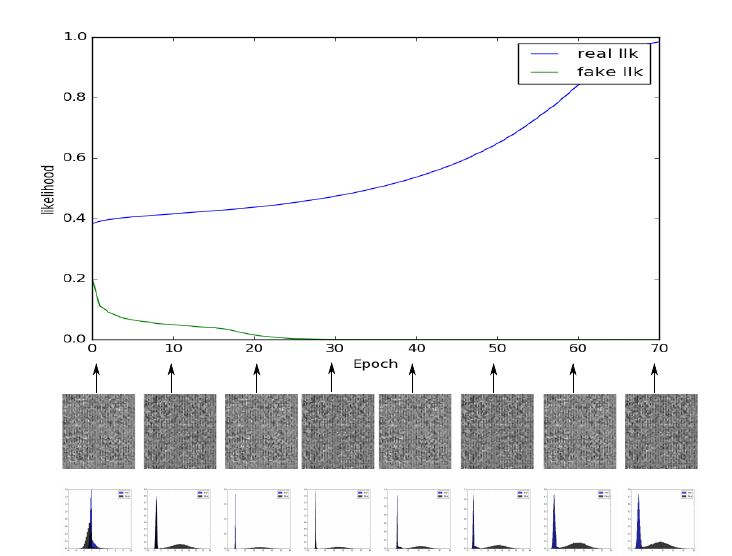}
  \caption{Generator is fixed and only discriminator is trained. }
  \label{fig:gen_nottrained}
\end{subfigure}
  \caption{Top: Likelihood curve for real and fake images as the model trains. Below that: images generated by the model in their respective epoch.
    Bottom: histogram of the bottleneck activations in their respective epoch. Blue corresponds to real and black to fake ones.
    }
  \label{fig:llk_curve}
\end{figure}

Figure \ref{fig:llk_curve} (left) shows an example of the likelihood of the discriminator with samples generated in different epochs.
On the right, an example is provided where the generator is fixed from the beginning and therefore the quality of the generated images does not improve.
The example on the left shows that at the beginning the quality of the generate images is bad, and consequently the likelihood of the generated images is low.
In the lower row, the histograms of the bottleneck activations are also shown on different epochs, separately for real and for fake images.
At the beginning of the training, two different clusters can be observed, while when the generator converges, they become more and more similar.

Looking at the likelihood plot on the left, it can clearly be seen that the likelihood shows a strong correlation with the quality of the generated image.
As the image quality is very bad (random), the likelihood goes to zero, although the likelihood of the real images goes to 1.
Also the histogram of the bottleneck activations stays separated into two clusters as the training does not improve the quality of the generated images.

\begin{figure}[h]
\centering
\begin{subfigure}{.4\textwidth}
  \centering
  \includegraphics[width=0.9\linewidth]{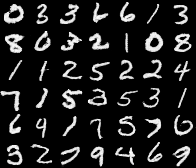}
  \caption{PGAN with 1-D bottleneck and \newline 1 Gaussian.}
  \label{fig:samples_pgan}
\end{subfigure}%
\begin{subfigure}{.4\textwidth}
  \centering
  \includegraphics[width=0.9\linewidth]{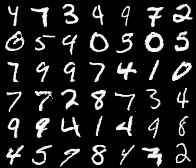}
  \caption{Least Squares GAN.\newline}
  \label{fig:samples_ls}
\end{subfigure}
\caption{Examples generated with a) PGAN, and b) LSGAN.}
\label{fig:generated_samples}
\end{figure}

Figure \ref{fig:generated_samples} shows some examples generated by LSGAN and PGAN.
It can be seen that, qualitatively, the images generated by PGAN with 1 Gaussian and 1-D bottleneck are quite similar to those of LSGAN.

\subsection{Stability of Training (1): Probability Landscape}
\label{subsec:boundary}

\begin{figure}[h]
\centering
\begin{subfigure}{.5\textwidth}
  \centering
  \includegraphics[width=0.9\linewidth]{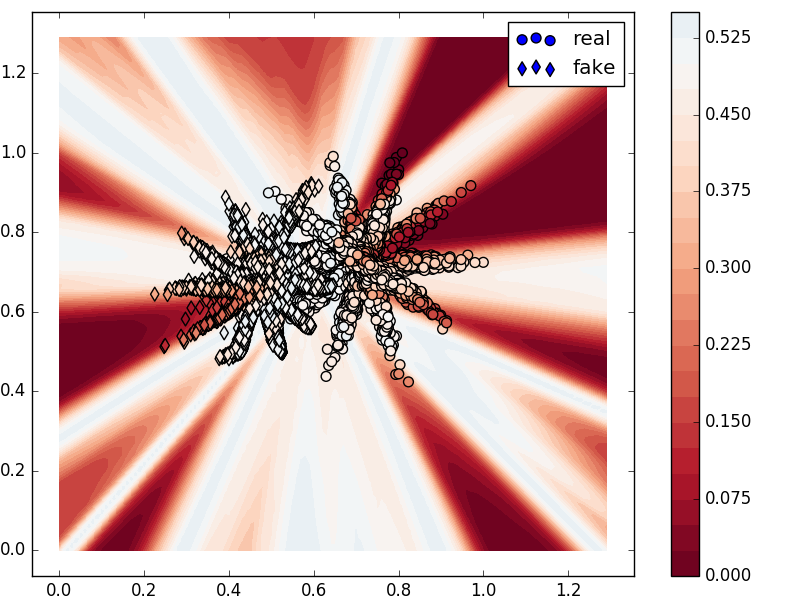}
  \caption{PGAN's discriminator probability landscape.}
  \label{fig:pgan_boundary}
\end{subfigure}%
\begin{subfigure}{.5\textwidth}
  \centering
  \includegraphics[width=0.9\linewidth]{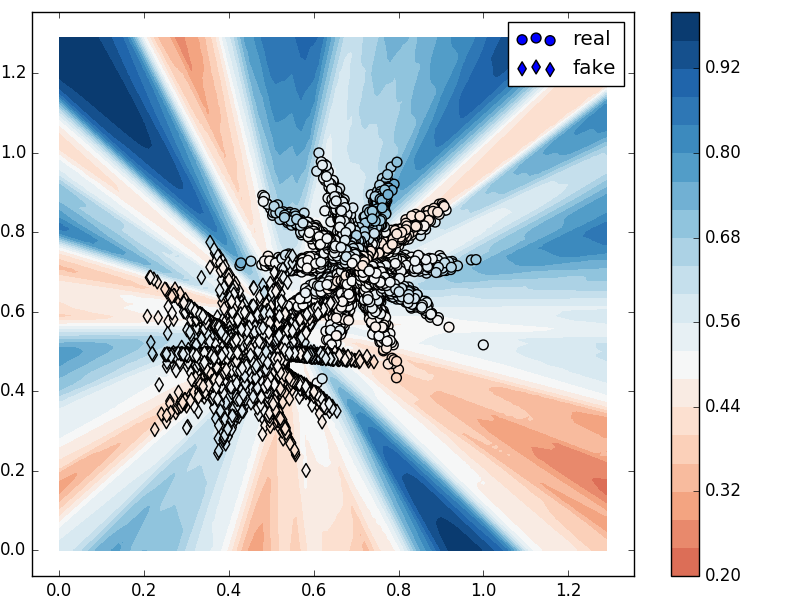}
  \caption{LSGAN's discriminator probability landscape.}
  \label{fig:lsgan_boundary}
\end{subfigure}
\caption{probability landscape of a 2-D space with 
LSGAN and PGAN discriminators, as well as the generated samples and the real samples. Color represents the probability (note the color bar): blue means higher and red means lower. Obviously, PGAN's discriminator assigns much lower probabilities to unseen parts of feature space than LSGAN.}
\label{fig:boundaries}
\end{figure}

As mentioned in \citep{nowozin2016f}, one cause for training instability in GANs is when there are many points in the domain of the discriminator with high assigned probability $P_{\mathit{real}}$ and low probability in $P_{\mathit{fake}}$.
This phenomenon is known as \emph{mode dropping}.
To look into this, we established a toy experiment using a 2D data generated by a 5-component GMM, where all components have the same mean, but a rotated covariance.
We used 2D data so that we can visualize the discriminator's probability assignment over feature space,
as well as the distribution of the real data and the generated data.
The result can be seen in Figure\ref{fig:boundaries}.
PGAN's discriminator (left) assigns much lower probabilities to many parts of data space, compared to the LSGAN.
This lower probability assignment does not affect negatively the ability of learning the distribution of the real data.
As can be seen in Figure\ref{fig:boundaries}, BGAN successfully learned the distribution of our toy example.

\subsection{Stability of Training (2): Gradients}
\label{subsec:boundary}

\citep{nowozin2016f} have shown that the norm of the gradients used to update the generator $G$
is important in stabilizing the training.
Furthermore they showed that when the generator is fixed and the discriminator is training, 
the gradients of the generator vanish or explode, if the generator has not converged.

\begin{figure}[htp]
\centering
\includegraphics[width=.3\textwidth]{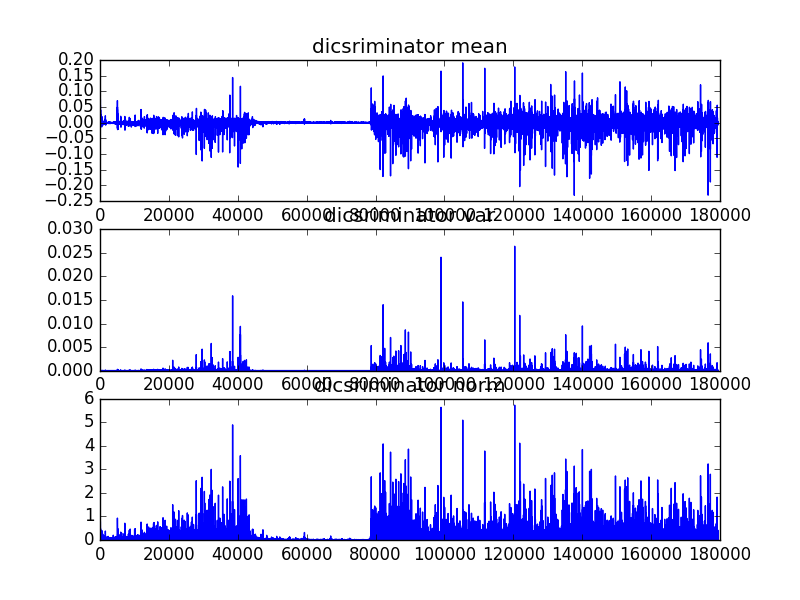}\quad
\includegraphics[width=.3\textwidth]{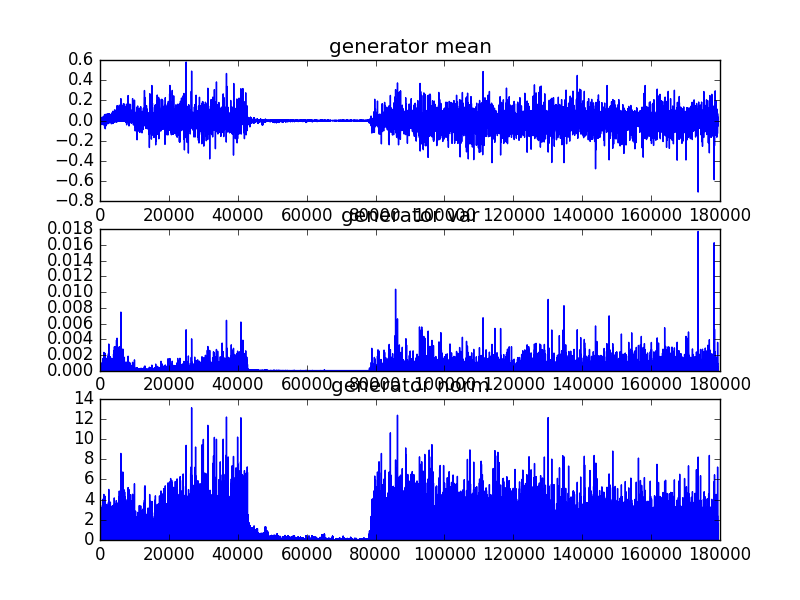}\quad
\includegraphics[width=.3\textwidth]{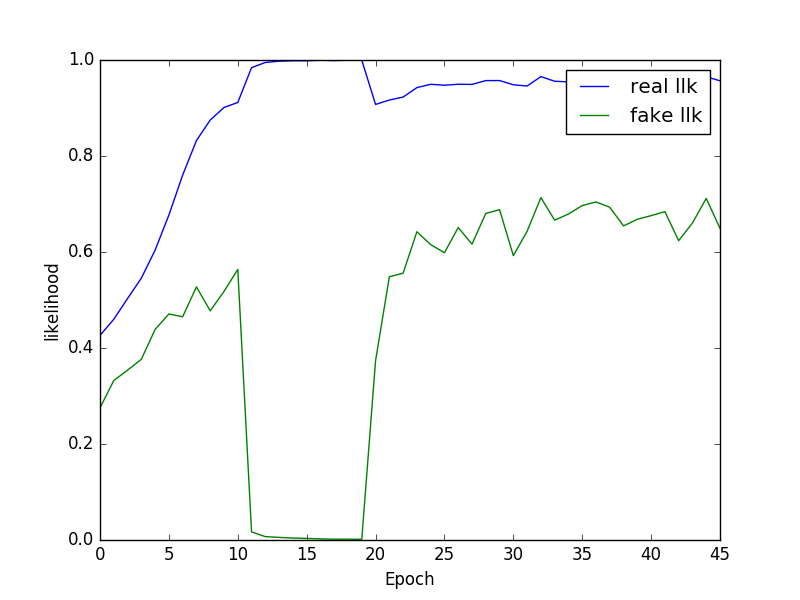}

\medskip

\includegraphics[width=.3\textwidth]{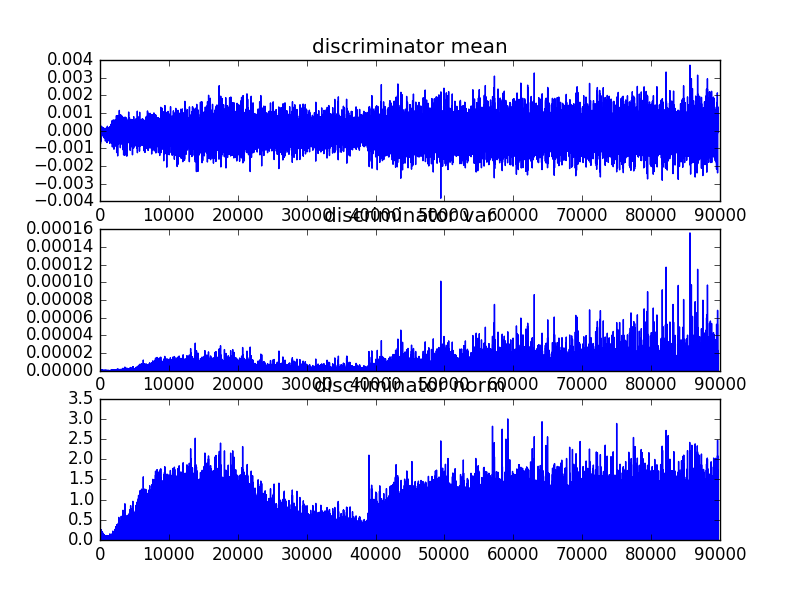}\quad
\includegraphics[width=.3\textwidth]{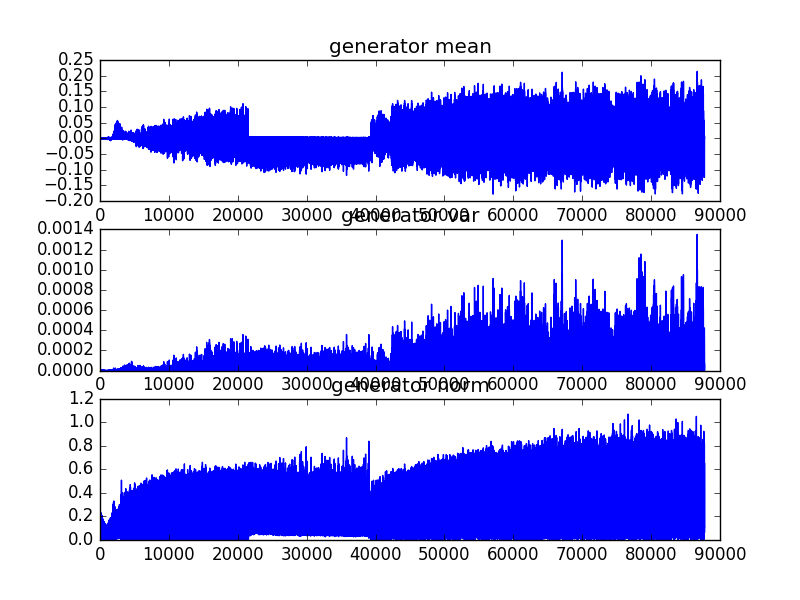}

\caption{Gradient On\textbackslash Off mechanism in PGAN: Gradient mean (first row), variance (second row) and norm (third row) of the discriminator and generator when the generator was blocked and released again.
On top left: PGAN discriminator, top middle: PGAN generator, top right PGAN likelihood.
On bottom left: LSGAN discriminator, bottom right: LSGAN generator.}
\label{fig:grads}
\end{figure}

Figure \ref{fig:grads} shows the results of an experiment where the generator is prevented from updating for 10 epochs, then again is being updated.
We visualize the gradient means, variances and norms for both generator and discriminator, for LSGAN and PGAN.
In can be seen that while the gradients in the generator and the discriminator still change in the LSGAN, 
PGAN uses an on\textbackslash off mechanism such that if the quality of the generated images does not significantly change, the gradients become quite small, with low norm and with almost no variation.
Another noticeable difference is the gradients of the discriminator.
In PGAN, when the generator does not change, the encoder in the discriminator does not need to change since the projection already maximizes the likelihood.
Therefore the new likelihood function is almost the same as the previous one.
Hence, the error is rather small which results in small and low-variance gradients produced by the discriminator.
Note that this is different from the vanishing the gradient problem, since the gradients are back to normal once the generator updates and changes the quality of the generated images.

The cause of this behavior is in the PGAN's discriminator.
Instead of using a classifier-like discriminator that needs both real and fake examples, 
PGAN uses a GMM to estimate a probability for the real data. 
Therefore, if the generator creates bad images, or doesn't improve, that does not significantly effect PGAN's discriminator.

\subsection{Stability of Training (3): The perfect discriminator problem}
\label{subsec:perfect_discriminator_problem}

One of the problems identified in \citep{arjovsky2017towards} is the \emph{perfect discriminator problem}.
When the discriminator gets closer to a perfect discrimination, the loss of the discriminator goes to zero and consequently
the gradient of the generator vanishes, effectively making it
impossible for the generator to continue improving.

We believe that the PGAN discriminator will have very low chance of becoming a perfect discriminator.
In fitting its Gaussian model, it focuses only on embeddings from the real data, skipping the fake inputs. Therefore the chances of the two being perfectly separated is quite low,
considering our loss pushes the fake embeddings to the side of the Gaussian bell which in both sides goes to the infinity.
Since our likelihood is continuous, the two embeddings will overlap in the tails.
Moreover, the amount of loss calculated to punish fake embeddings further and further from the mean of the real embeddings 
can not be very large since the Gaussian likelihood decreases exponentially when moving away from the mean.
Therefore, it is very unlikely for the discriminator to completely separate the two.

\begin{figure}
 \centering
 \begin{subfigure}{.5\textwidth}
   \centering
   \includegraphics[width=1.\linewidth]{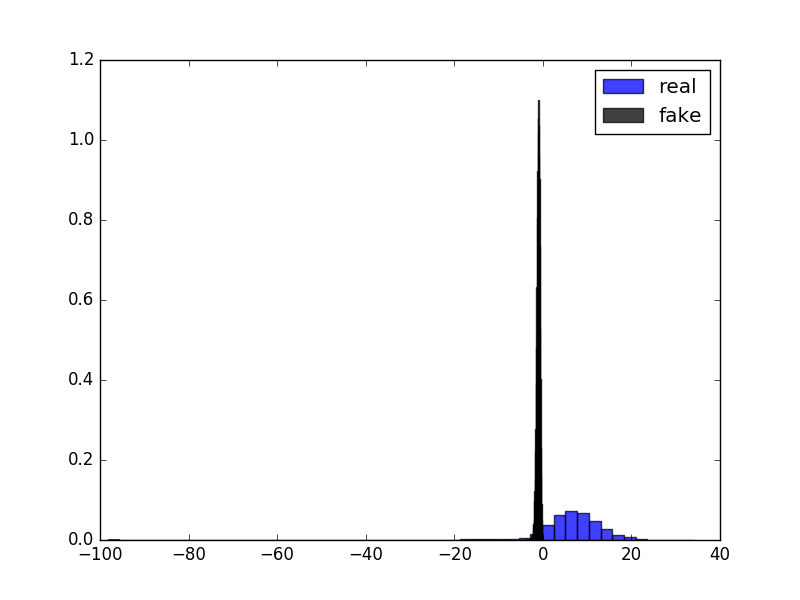}
   \caption{LSGAN.}
   \label{fig:binary_clf}
 \end{subfigure}%
 \begin{subfigure}{.5\textwidth}
   \centering
   \includegraphics[width=1.\linewidth]{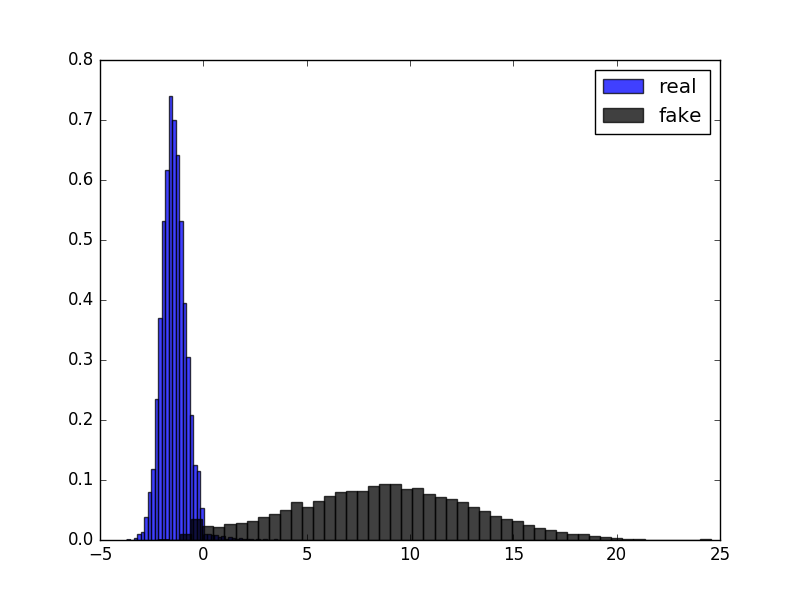}
   \caption{PGAN.}
   \label{fig:gm_clf}
 \end{subfigure}
 \caption{Histograms of activations of the bottleneck layer in LSGAN and PGAN. (Remember that in our experiments, the bottleneck layer
 consists of a single unit.)
 To train the discriminator to its best, the generator was fixed from the beginning.}
 \label{fig:perfectd}
 \end{figure}

This can also be seen in Figure \ref{fig:perfectd}.
To show that how separate the two embeddings can be when the discriminator gets better and better, 
we provide the histograms of real and fake embeddings in LSGAN and PGAN when the generator is kept fixed.
For the LSGAN case on the left, we can see that the very sharp spikes of the fake embeddings permit a near-perfect separation.
For PGAN the fake embedding distribution is extended along the axis and it is very unlikely for the two class to be perfectly separated.

Another fact that \cite{arjovsky2017towards} pointed out is the benefits of
using a discriminator with a \emph{softer measure}.
They suggest that incorporating some notion of \emph{distance} between elements in the support of $P_{\mathit{real}}$ and support of $P_{\mathit{fake}}$ will help to tackle the perfect discriminator problem.
The likelihood of our discriminator can be seen as a notion of distance between elements in the support of $P_{\mathit{real}}$ and $P_{\mathit{fake}}$ in the following way:
if one data point from $\mathbb{P}_{\mathit{fake}}$ is close to data points in $\mathbb{P}_{\mathit{real}}$, then the Gaussian likelihood assigned to it will be higher.
Also if data points in $\mathbb{P}_{\mathit{fake}}$ are far from data points in $\mathbb{P}_{\mathit{real}}$, for those data points the likelihood will be low.

\section{Conclusion}
\label{sec:conclusion}
In this paper, we introduced a new variant of GANs with an integrated probabilistic model, 
namely the Probabilistic Generative Adversarial Network.
PGAN takes advantage of a Gaussian Mixture Model to assign Gaussian likelihoods to the embedded images.
This likelihood is used for optimizing both the discriminator and the generator.
One benefit of PGAN is that this likelihood has a strong correlation with the divergence of the GAN and the quality of the generated images and therefore can be used as a measure of training.
We also provided some evidence that PGAN is better able to cope with instability problems in the GAN training procedure.
We showed empirical results of the more stable training by using the tools proposed for identifying the sources of instability in GANs.


\small
\bibliographystyle{plainnat}
\bibliography{refs}

\begin{thebibliography}{9}
\providecommand{\natexlab}[1]{#1}
\providecommand{\url}[1]{\texttt{#1}}
\expandafter\ifx\csname urlstyle\endcsname\relax
  \providecommand{\doi}[1]{doi: #1}\else
  \providecommand{\doi}{doi: \begingroup \urlstyle{rm}\Url}\fi

\bibitem[Arjovsky and Bottou(2017)]{arjovsky2017towards}
Martin Arjovsky and L{\'e}on Bottou.
\newblock Towards principled methods for training generative adversarial
  networks.
\newblock In \emph{NIPS 2016 Workshop on Adversarial Training. In review for
  ICLR}, volume 2016, 2017.

\bibitem[Arjovsky et~al.(2017)Arjovsky, Chintala, and
  Bottou]{arjovsky2017wasserstein}
Martin Arjovsky, Soumith Chintala, and L{\'e}on Bottou.
\newblock Wasserstein gan.
\newblock \emph{arXiv preprint arXiv:1701.07875}, 2017.

\bibitem[Goodfellow et~al.(2014)Goodfellow, Pouget-Abadie, Mirza, Xu,
  Warde-Farley, Ozair, Courville, and Bengio]{goodfellow2014generative}
Ian Goodfellow, Jean Pouget-Abadie, Mehdi Mirza, Bing Xu, David Warde-Farley,
  Sherjil Ozair, Aaron Courville, and Yoshua Bengio.
\newblock Generative adversarial nets.
\newblock In \emph{Advances in neural information processing systems}, pages
  2672--2680, 2014.

\bibitem[Kingma and Ba(2014)]{kingma2014adam}
Diederik Kingma and Jimmy Ba.
\newblock Adam: A method for stochastic optimization.
\newblock \emph{arXiv preprint arXiv:1412.6980}, 2014.

\bibitem[Mao et~al.(2016)Mao, Li, Xie, Lau, Wang, and Smolley]{mao2016least}
Xudong Mao, Qing Li, Haoran Xie, Raymond~YK Lau, Zhen Wang, and Stephen~Paul
  Smolley.
\newblock Least squares generative adversarial networks.
\newblock \emph{arXiv preprint ArXiv:1611.04076}, 2016.

\bibitem[Nowozin et~al.(2016)Nowozin, Cseke, and Tomioka]{nowozin2016f}
Sebastian Nowozin, Botond Cseke, and Ryota Tomioka.
\newblock f-gan: Training generative neural samplers using variational
  divergence minimization.
\newblock In \emph{Advances in Neural Information Processing Systems}, pages
  271--279, 2016.

\bibitem[Radford et~al.(2015)Radford, Metz, and
  Chintala]{radford2015unsupervised}
Alec Radford, Luke Metz, and Soumith Chintala.
\newblock Unsupervised representation learning with deep convolutional
  generative adversarial networks.
\newblock \emph{arXiv preprint arXiv:1511.06434}, 2015.

\bibitem[Shalev-Shwartz and Ben-David(2014)]{shalev2014understanding}
Shai Shalev-Shwartz and Shai Ben-David.
\newblock \emph{Understanding machine learning: From theory to algorithms}.
\newblock Cambridge university press, 2014.

\bibitem[Zhao et~al.(2016)Zhao, Mathieu, and LeCun]{zhao2016energy}
Junbo Zhao, Michael Mathieu, and Yann LeCun.
\newblock Energy-based generative adversarial network.
\newblock \emph{arXiv preprint arXiv:1609.03126}, 2016.

\end{thebibliography}

\newpage
\appendix
\section{Appendices}
\subsection{Network architectures}
\label{ap:a}

\begin{table}[h]
  \caption{The architectures used in PGAN. The LSGAN uses the same generator. The LSGAN discriminator is the same as in b, but with a 1-dim bottleneck and no GMM.}
  \label{tab:architectures}
  \centering
\begin{subfigure}{.5\textwidth}
  \centering
    \begin{tabular}{l}
	Input $1 \times 100$\\
  	\hline
	Dense Layer (1024, LReLu\footnote{leaky rectified linear units.}) + BN\footnote{Batch normalization layer.}\\
	Dense Layer (6272, LReLu) + BN\\
	Reshape Layer ($1 \times 128 \times 7 \times 7$)\\
  	\hline
    Deconv Layer (64, $5 \times 5$) + BN\\
    Deconv Layer (1, $5 \times 5$, stride=2) + BN\\
	\hline
   	Nonlinearity (Sigmoid)\\
  \end{tabular}
  \caption{The generator's architecture for MNIST.}
  \label{fig:gen_arch}
\end{subfigure}%
\begin{subfigure}{.5\textwidth}
  \centering
    \begin{tabular}{l}
    Input $1 \times 28 \times 28$\\
  	\hline
	Conv Layer (64, $5 \times 5$, stride=2, LReLu) + BN\\
	Conv Layer (128, $5 \times 5$, stride=2, LReLu) + BN\\
	\hline
	Dense Layer (1024, LReLu) + BN 	(bottleneck) \\
    Dense Layer (bottleneck size, LReLu) + BN 	(bottleneck) \\
	\hline
	Gaussian Mixture Model (Diagonal Covariance)
  \end{tabular}
  \caption{The discriminator's Architecture for MNIST}
  \label{fig:disc_arch}
\end{subfigure}  
\end{table}

\begin{table}[h]
  \caption{The architectures used in PGAN.}
  \label{tab:architectures_cifar}
  \centering
\begin{subfigure}{\textwidth}
  \centering
    \begin{tabular}{l}
	Input $1 \times 100$\\
  	\hline
     TransConv Layer (256, $4 \times 4$, stride=1,crop=0,nonlin=LReLu(0.2),init=Normal(0.02, 0))+BN\\
	TransConv Layer (128, $4 \times 4$, stride=2, crop=1,nonlin=LReLu(0.2),init=Normal(0.02, 0))+BN\\
	TransConv Layer (64, $4 \times 4$, stride=2, crop=1,nonlin=LReLu(0.2),init=Normal(0.02, 0))+BN\\
	TransConv Layer (3, $4 \times 4$, stride=2, crop=1,nonlin=Sigmoid,init=Normal(0.02, 0))+BN\\
	\hline
   	Nonlinearity (Sigmoid)\\
  \end{tabular}
  \caption{The generator's architecture for CIFAR-10.}
  \label{fig:gen_arch}
\end{subfigure}
\begin{subfigure}{\textwidth}
  \centering
    \begin{tabular}{l}
    Input $1 \times 28 \times 28$\\
  	\hline
	Conv Layer (64, $4 \times 4$, stride=2, pad=1,nonlin=LReLu(0.2),init=Normal(0.02, 0))
\\
Conv Layer (128, $4 \times 4$, stride=2, pad=1,LReLu(0.2),init=Normal(0.02, 0))+BN
\\
Conv Layer (256, $2 \times 2$, stride=2, pad=1,LReLu(0.2),init=Normal(0.02, 0))+BN
\\
Conv Layer (1, $4 \times 4$, stride=2, pad=1,LReLu(0.2)\\
	\hline
	Dense Layer (1024, LReLu) + BN 	(bottleneck) \\
    Dense Layer (bottleneck size, LReLu) + BN 	(bottleneck) \\
	\hline
	Gaussian Mixture Model (Diagonal Covariance)
  \end{tabular}
  \caption{The discriminator's Architecture for CIFAR-10}
  \label{fig:disc_arch}
\end{subfigure}  
\end{table}

\subsection{Extended empirical results}
\label{ap:b}
\begin{figure}[h]
  \centering
  \includegraphics[width=1.\linewidth]{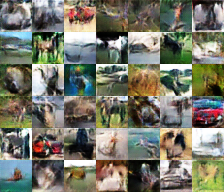}
  \caption{Generated images by PGAN with one Gaussian and 1-D bottleneck from CIFAR-10.}
\end{figure}

\begin{figure}[h]
  \centering
  \includegraphics[width=1.\linewidth]{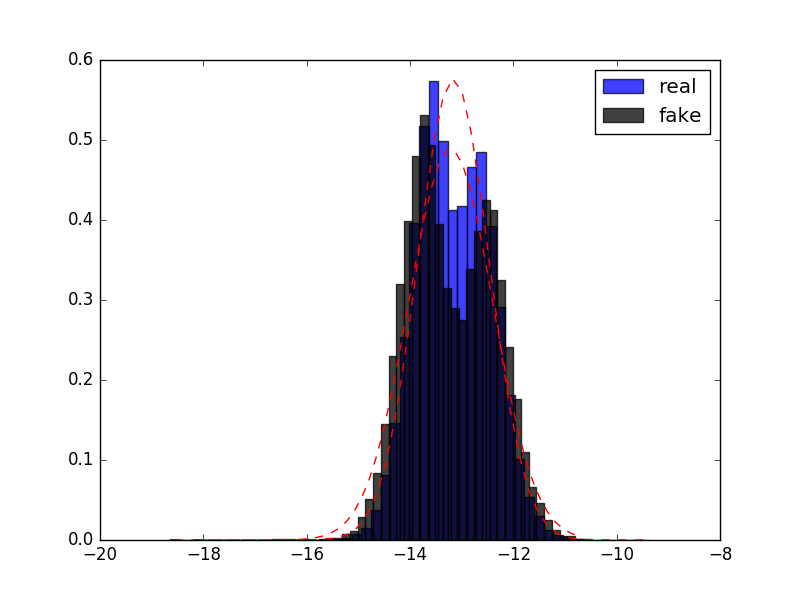}
  \caption{Histogram of bottleneck for a PGAN with one Gaussian and a 1-D bottleneck trained on CIFAR-10.}
\end{figure}

\begin{figure}[h]
  \centering
  \includegraphics[width=1.\linewidth]{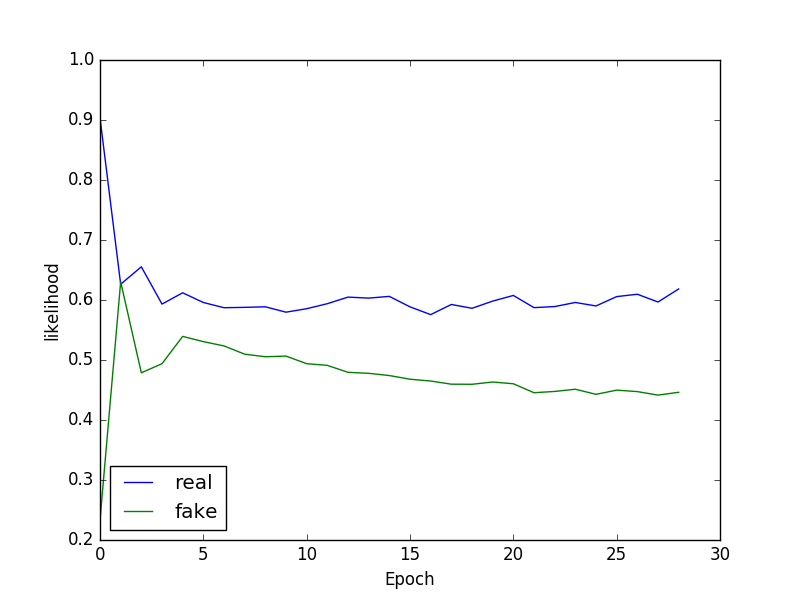}
  \caption{Likelihoods of the real and fake images for CIFAR-10.}
\end{figure}

\begin{figure}[h]
  \centering
  \includegraphics[width=1.\linewidth]{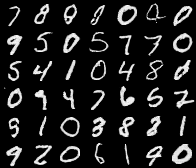}
  \caption{Generated images by PGAN with one Gaussian and 1-D bottleneck from MNIST.}
\end{figure}

\begin{figure}[h]
  \centering
  \includegraphics[width=1.\linewidth]{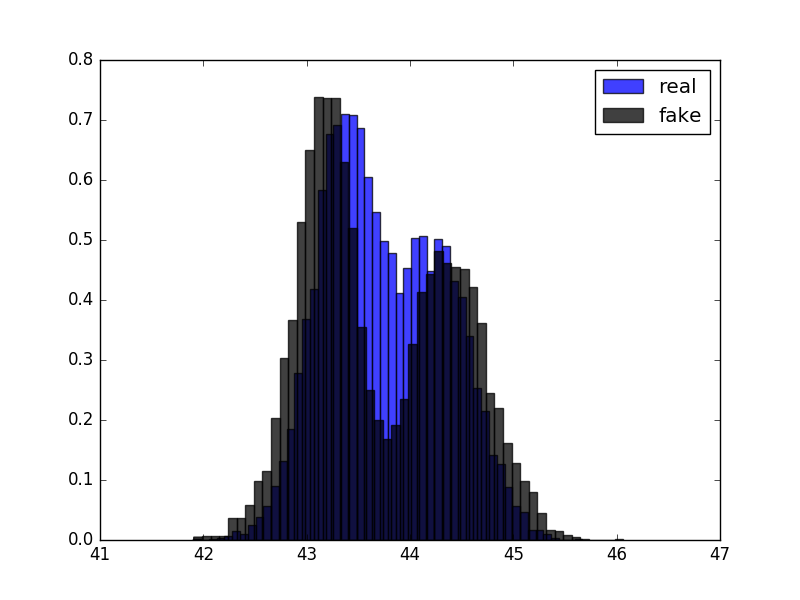}
  \caption{Histogram of bottleneck for a PGAN with one Gaussian and a 1-D bottleneck trained on MNIST.}
\end{figure}
\begin{figure}[h]
  \centering
  \includegraphics[width=1.\linewidth]{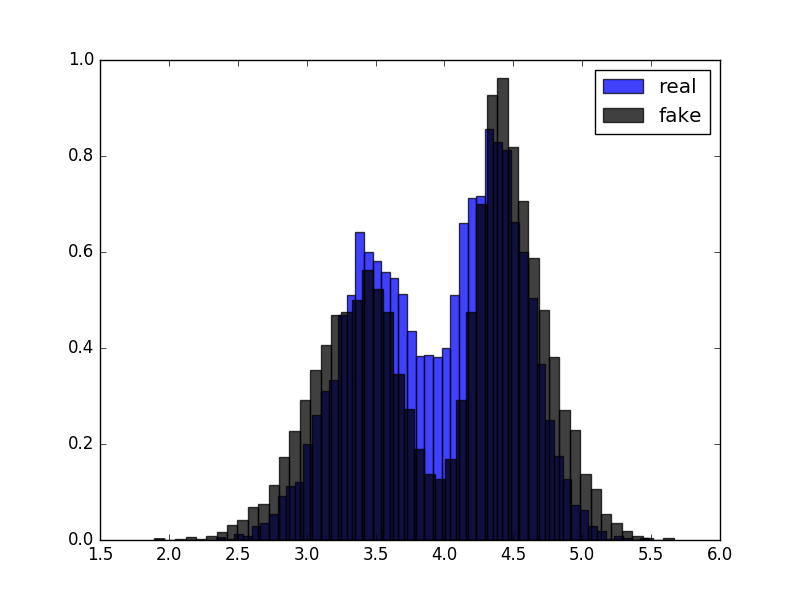}
  \caption{Histogram of bottleneck for a PGAN with two Gaussian components and a 1-D bottleneck trained on MNIST.}
\end{figure}

\begin{figure}[h]
  \centering
  \includegraphics[width=1.\linewidth]{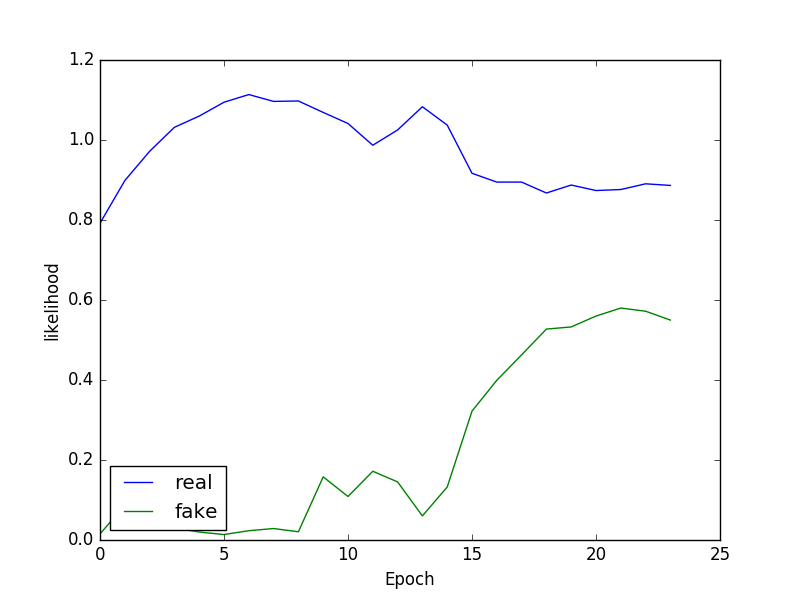}
  \caption{Likelihoods of the real and fake images for MNIST.}
\end{figure}

\begin{figure}[h]
  \centering
  \includegraphics[width=1.\linewidth]{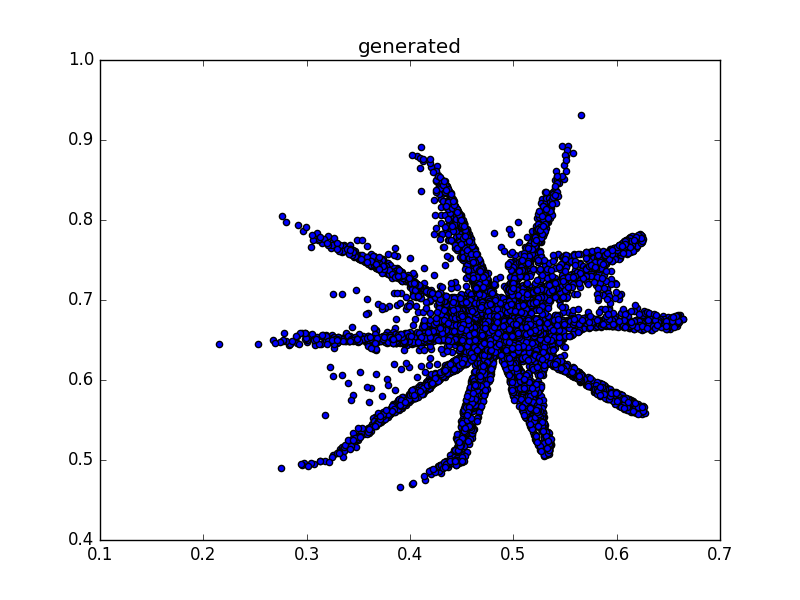}
 \caption{Generated samples by PGAN with one Gaussian and 1-D bottleneck from our toy example.}
  \label{fig:generated_toyd}
\end{figure}
\begin{figure}[h]
  \centering
  \includegraphics[width=1.\linewidth]{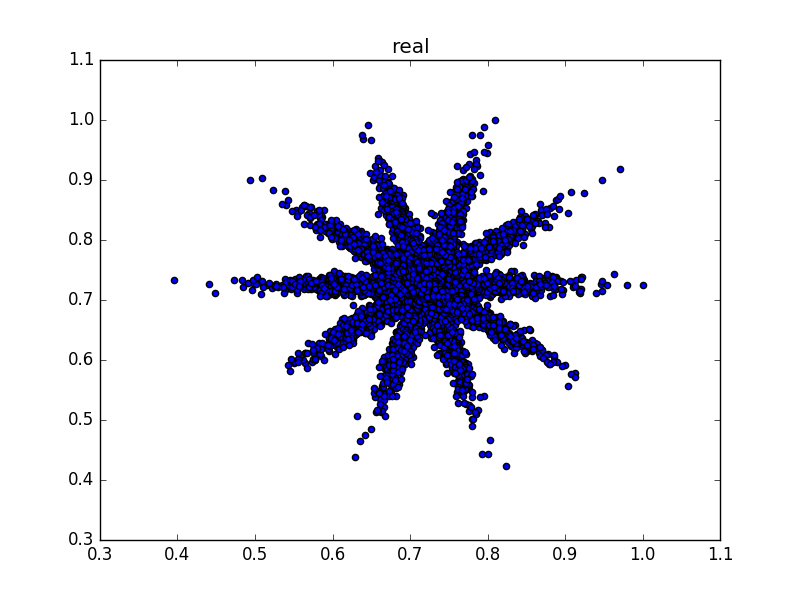}
  \caption{Real samples from of our 2-D toy example.
  A 5-mixtures GMM with an equal mean and a rotated covariance was used to generate this toy example.}
  \label{fig:toyd}
\end{figure}

\end{document}